\documentclass{article}
\usepackage{ijcai22}

\usepackage{times}

\usepackage{soul}
\usepackage{url}
\usepackage[hidelinks]{hyperref}
\usepackage[utf8]{inputenc}
\usepackage[small]{caption}
\usepackage{graphicx}
\usepackage{amsmath}
\usepackage{booktabs}
\usepackage{amsfonts,amsmath,amsthm}
\usepackage{tikz,pgffor}
\usepackage{ifthen}
\usetikzlibrary{arrows,backgrounds,calc}
\usetikzlibrary{automata,positioning,arrows,shapes,math,arrows.meta,decorations.pathmorphing}
\tikzstyle{min}=[thick,circle,draw,minimum size=1.4em,inner sep=0em,text centered]
\usepackage{scalerel}
\usepackage{algorithm}
\usepackage{enumitem}
\usepackage[noend]{algpseudocode}
\newcommand{\commentsymbol}{//}
\algrenewcommand\algorithmiccomment[1]{\hfill\textcolor{ourblue}{\commentsymbol{} #1}}
\newcommand{\LineComment}[2][1]{\Statex \hspace{#1\dimexpr\algorithmicindent}\textcolor{ourblue}{\commentsymbol{} #2}}
\usepackage{flushend}

\newcommand{\Nset}{\mathbb{N}}
\newcommand{\Dist}{\operatorname{Dist}}
\newcommand{\M}{\mathcal{M}}
\newcommand{\F}{\mathbb{F}}
\newcommand{\tm}{\operatorname{tm}}
\newcommand{\Mem}{\operatorname{M}}
\newcommand{\conf}{\operatorname{Conf}}
\newcommand{\Tm}{\mathbb{T}}
\newcommand{\dhat}[1]{\smash{\hat{#1}}}
\newcommand{\Ac}{\mathcal{A}_{c}}
\newcommand{\Ae}{\mathcal{A}_{e}}
\newcommand{\Out}{\operatorname{Out}}

\let\hat\widehat

\definecolor{ourblue} {RGB}{60,119,177}
\definecolor{ourorange} {RGB}{236,134,54}
\hypersetup{colorlinks=true, linkcolor=ourblue, citecolor=ourorange, urlcolor=ourblue}

\theoremstyle{definition}
\newtheorem{example}{Example}

\usepackage{xspace}
\makeatletter
\DeclareRobustCommand\onedot{\futurelet\@let@token\@onedot}
\def\@onedot{\ifx\@let@token.\else.\null\fi\xspace}

\newcommand\eg{{e.g}\onedot} 
\newcommand\ie{{i.e}\onedot} 
\newcommand\wrt{w.r.t\onedot} 
\makeatother

\title{General Optimization Framework for Recurrent Reachability Objectives}

\author{
David Kla\v ska\and
Anton\'{\i}n Ku\v cera\footnote{contact author \textsl{$\langle$tony@fi.muni.cz$\rangle$}}\and
V\'{\i}t Musil\And
Vojt\v ech \v Reh\'ak\\
\affiliations
Faculty of Informatics, Masaryk University, Brno, Czech Republic\\
\emails
}

\begin{document}

\maketitle

\begin{abstract}
We consider the mobile robot path planning problem for a class of recurrent reachability objectives.
These objectives are parameterized by the expected time needed to visit one position from another, the expected square of this time, and also the frequency of moves between two neighboring locations.
We design an efficient strategy synthesis algorithm for recurrent reachability objectives and demonstrate its functionality on non-trivial instances.
\end{abstract}

\section{Introduction}
\label{sec-intro}

In mobile robot path planning, the terrain is represented as a directed graph where the vertices are robot positions, the edges correspond to possible robot moves, and every edge is assigned the corresponding \emph{traversal time}. 
A \emph{moving strategy} specifies how the robot moves from vertex to vertex, and it can be \emph{deterministic} or \emph{randomized}.  

In this work, we concentrate on \emph{infinite-horizon} path planning problems where the robot performs a recurring task such as surveillance or periodic maintenance. The standard tool for specifying infinite-horizon objectives are \emph{frequency-based} objective functions parameterized by the \emph{limit frequency} of visits to the vertices.  
Unfortunately, these functions are \emph{insufficient} for expressing subtle optimization criteria used in specific areas such as robotic patrolling, and cannot faithfully capture all crucial properties of randomized strategies such as deviations/variances of relevant random variables. The latter deficiency represents a major problem often resolved indirectly by considering only \emph{deterministic} strategies, even in scenarios where randomization achieves strictly better performance and is easy to implement (see Example~\ref{exa-random}). 
\smallskip

\paragraph{Our contribution.}
We design and investigate a class of \emph{recurrent reachability objective functions} based on the following parameters: 
\begin{enumerate}[label={\rm (\arabic*)}]
\item\label{en:1} the limit frequency of edges;
\item\label{en:2} the expected time to hit a given set of vertices from another given vertex;
\item\label{en:3} the expected \emph{square} of the time to hit a given set of vertices from another given vertex.   
\end{enumerate}
Note that using~\ref{en:1}, one can express the frequency of visits to vertices, and \ref{en:2} and~\ref{en:3} allow to express the \emph{variance} and \emph{standard deviation} of the time to hit a given set of vertices from another given vertex.
Thus, the recurrent reachability objective functions can ``punish'' large deviations from the expected values, allowing for \emph{balancing performance with stochastic stability}.

Computing an optimal moving strategy for a given recurrent reachability objective is computationally hard. One can easily reduce the $\mathsf{NP}$-hard Hamiltonian cycle problem to the problem of deciding whether the minimum of a certain recurrent reachability objective function is bounded by a given constant.
This means there is no efficient strategy synthesis algorithm with optimality guarantees unless $\mathsf{P=NP}$.

We design a strategy synthesis algorithm based on gradient descent applicable to \emph{arbitrary recurrent reachability objectives involving piecewise differentiable continuous functions}.
The algorithm efficiently computes a \emph{finite-memory} randomized strategy where the memory is used to ``remember'' some relevant information about the history of visited vertices. Although the (sub)optimality of this strategy is not guaranteed for the reasons mentioned above, our experiments show that the algorithm can solve instances requiring non-trivial insights and produce solutions close to theoretical optima.

Thus, we obtain a \emph{general and efficient optimization framework for an expressively rich class of non-linear infinite-horizon objectives capable of solving problems beyond the reach of existing methods}.

\subsection{Motivating Example}
In this section, we give an example illustrating the limitations of frequency-based objectives and deterministic strategies, and we show how randomization and recurrent reachability objectives help to overcome these problems.

In robotic patrolling, some vertices in the terrain graph are declared as \emph{targets}, and the robot aims to discover possible intrusions at the targets. One standard measure for the protection achieved by a given moving strategy is the maximal \emph{average idleness} of a target \cite{HZHH:multi-robot-patrol-survey,ARSTMCC:multi-patrolling-survey,PR:multi-patrolling-survey}. 
In the language of Markov chains, this corresponds to the maximal \emph{renewal time} of a target, \ie, $\max_{\tau \in T} 1/f_\tau$, where $T$ is the set of all targets and $f_\tau$ is the frequency of visits to~$\tau$ (recall that $1/f_\tau$ is the expected time of revisiting $\tau$).

Existing works about minimizing idleness aim at constructing a \emph{deterministic} moving strategy, \ie, a cycle in the underlying graph \cite{HZHH:multi-robot-patrol-survey,ARSTMCC:multi-patrolling-survey,PR:multi-patrolling-survey}. The next example shows that using \emph{randomized} strategies brings additional benefits that have not been exploited so far.

\begin{figure}[t]\centering
\begin{tikzpicture}[x=2cm, y=1.1cm, scale=0.7,font=\small]
\foreach \x/\c/\l in {0/0/c,1/2/b,2/3.5/a}{%
    \coordinate (a) at (0,\c);
    \node at ($(a) +(-.8,0)$) {(\l)};
    \node [min,double] (T1\x) at (a) {$\tau_1$};
    \node [min,double] (T2\x) at ($ (a) + (3,0) $) {$\tau_2$};
    \node [min] (V1\x) at ($ (a) + (1,0) $) {$v_1$};
    \node [min] (V2\x) at ($ (a) + (2,0) $) {$v_2$};
    \ifthenelse{\x=2}{%
    \draw[stealth-stealth,very thick] (T1\x) -- node[above] {$ $}  (V1\x);
    \draw[stealth-stealth,very thick] (V1\x) -- node[above] {$ $} (V2\x);
    \draw[stealth-stealth,very thick] (V2\x) -- node[above] {$ $} (T2\x);
    }{%
    \draw[stealth-stealth,very thick] (T1\x) -- (V1\x);
    \draw[stealth-stealth,very thick] (V1\x) -- (V2\x);
    \draw[stealth-stealth,very thick] (V2\x) -- (T2\x);
    \draw[-stealth,rounded corners] (T1\x) -- ($(T1\x) +(0,-.5)$) -- ($(V1\x) + (0,-.5)$);  
    \draw[rounded corners] ($(V1\x) +(0,.5)$) -| (T1\x);
    \draw[-stealth,rounded corners] (T2\x) -- ($(T2\x) +(0,.5)$) -- ($(V2\x) + (0,.5)$);
    \draw[rounded corners] ($(V2\x) +(0,-.5)$) -| (T2\x);  
    }
    \ifthenelse{\x=1}{%
    \draw[-stealth,rounded corners] ($(V2\x) +(0,.5)$) -- ($(V1\x) +(0,.5)$); 
    \draw[-stealth,rounded corners] ($(V1\x) +(0,-.5)$) -- ($(V2\x) +(0,-.5)$); 
    }{} 
    \ifthenelse{\x=0}{%
    \draw[-stealth, densely dashed, rounded corners] ($(V2\x) +(0,.5)$) -- node[above]{$1-x$} ($(V1\x) +(0,.5)$); 
    \draw[-stealth, densely dashed, rounded corners] ($(V1\x) +(0,-.5)$) -- node[below]{$1-x$} ($(V2\x) +(0,-.5)$); 
    \draw[-stealth, densely dashed, rounded corners] ($(V2\x) +(0,.5)$) -- ($(V2\x) +(0,.8)$) --
       node[above]{$x$}  ($(T2\x) +(0.05,.8)$) -- ($(T2\x) + (0.05,0.35)$);    
    \draw[-stealth, densely dashed, rounded corners] ($(V1\x) +(0,-.5)$) -- ($(V1\x) +(0,-.8)$) --
       node[below]{$x$}  ($(T1\x) +(-.05,-.8)$) -- ($(T1\x) - (0.05,0.35)$);    
    }{}
}
\end{tikzpicture}
\caption{On graph~(a) with targets $\tau_1$ and $\tau_2$, randomized strategy (c) has lower \emph{renewal time} than the deterministic strategy (b).}
\label{fig-rand}
\end{figure}
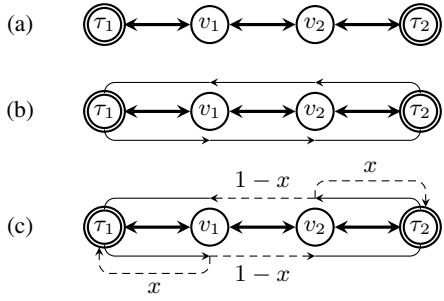

\begin{example}
\label{exa-random}
Consider the graph of Fig.~\ref{fig-rand}a with two targets $\tau_1,\tau_2$ where traversing every edge takes one time unit. Let $\sigma_b$ be a \emph{deterministic} strategy alternately visiting $\tau_1$ and $\tau_2$, see Fig.~\ref{fig-rand}b. Then, both $\tau_1$ and $\tau_2$ are revisited in $6$ time units, and the maximal renewal time is~$6$.

At first glance, it seems the robot cannot do any better. However, consider the randomized strategy $\sigma_c$ of Fig.~\ref{fig-rand}c. When the robot comes to $v_1$ from $\tau_1$, it returns to $\tau_1$ with probability~$x$. With the remaining probability $1-x$, it continues to $v_2$. A symmetric decision is taken when the robot comes to $v_2$ from $\tau_2$. 

As $x \rightarrow 1$ in $\sigma_c$, the frequency of visits to $\tau_1,\tau_2$ approaches $1/4$, \ie, the renewal time of $\tau_1,\tau_2$ approaches~$4$. However, pushing $x$ close to $1$ results a strategy where the robot needs \emph{very long time} to move from $\tau_1$ to $\tau_2$ (and vice versa). Such a strategy is clearly \emph{not appropriate} for surveillance purposes. So, we may refine the objective and minimize the maximum of renewal times \emph{and} the expected time of visiting one target from the other (note that this recurrent reachability objective is \emph{not} frequency-based).  A simple computation reveals that the \emph{optimal} choice is then setting $x =  (5{-}\sqrt{17})/4$, yielding the maximum $\approx 5.56$. This strategy does not have the above defect and \emph{outperforms} the deterministic strategy $\sigma_b$.

Another way of eliminating the ``defect'' of $\sigma_c$ for $x \rightarrow 1$ is to control the \emph{variance} of the renewal time of $\tau_1$, $\tau_2$, which approaches $\infty$ as $x \rightarrow 1$.
Using recurrent reachability objectives, this can be expressed as, \eg, minimizing a weighted sum of the maximal renewal time and the maximal variance of renewal time. Thus, one may \emph{trade} the value of renewal time with its stochastic stability. 
\end{example}

\subsection{Related Work}

The \emph{finite-horizon} path planning problem involving finding a feasible path between two given positions is one of the most researched subject in mobile robotics (see, \eg, \cite{Choset:book,LaValle:book}). Recent technological advances motivate the study of \emph{infinite-horizon} path planning problems where the robot performs an uninterrupted task such as 
persistent data-gathering \cite{STBR:temporal-planning-IJRS} or patrolling \cite{HZHH:multi-robot-patrol-survey,ARSTMCC:multi-patrolling-survey,PR:multi-patrolling-survey}. The (classical) vehicle routing problem and the generalized traveling salesman problem \cite{TV:book} can also be seen as infinite-horizon path planning problems. The constructed strategies were originally meant for humans (drivers, police squads, etc.) and hence they are \emph{deterministic}. The only exception is adversarial patrolling based on Stackelberg equilibria \cite{SFAKT:Stackelberg-Security-Games,YKKCT:Stackelberg-Nash-security} where randomization was soon identified as a crucial tool for decreasing the patroller's predictability.

The existing objectives studied in infinite-horizon path planning problems are mostly based on long-run average quantities related to vertex frequencies or cycle lengths, such as mean payoff or renewal time (see, \eg, \cite{Puterman:book}). General-purpose specification languages for infinite-horizon objectives are mostly based on linear temporal logics (see, \eg, \cite{PLGG:plans-LTL-IJCAI,WTM:trajectory-LTL-ICRA,USDBR:multi-robot-plan-TL-IJRS,BKV:motion-plan-temporal-CDC}). A formula of such a logic specifies desirable properties of the constructed path, and may include additional mechanisms for identifying a subset of optimal paths. The optimality criteria are typically based on distances between vertices along a path, and the constructed strategies are deterministic.      

To the best of our knowledge, the recurrent reachability objective functions represent the first general-purpose specification language allowing to \emph{utilize the benefits of randomized strategies and even specify tradeoffs between performance and stochastic stability}. Even in a broader context of stochastic programming, the existing works about balancing quantitative features of probabilistic strategies have so far been limited to some variants of mean payoff and limit frequencies \cite{BCFK:performance-stability-JCSS}. These results are not applicable in our setting due to a different technical setup. Furthermore, our specification language is not limited to frequency-based objectives.

\section{The Model}
\label{sec-model}

In the rest of this paper, we use $\Nset$ and $\Nset_+$ to denote the sets of non-negative and positive integers. The set of all probability distributions over a finite set $A$ is denoted by $\Dist(A)$. A~distribution $\nu$ over $A$ is \emph{positive} if $\nu(a) > 0$ for all $a \in A$, and \emph{Dirac} if $\nu(a) =1$ for some $a \in A$. 

A finite-state \emph{Markov chain} is a pair $\M = (S,P)$ where $S$~is a finite set of states and $P \colon S \times S \to [0,1]$ is a stochastic matrix where the sum of every row is equal to~$1$.  A \emph{bottom strongly connected component (BSCC)} of $\M$ is a maximal $B \subseteq S$ such that for all $s,t \in B$ we have that $s$ can reach $t$ with positive probability (\ie, $P^m(s,t) > 0$ for some \mbox{$m \in \Nset$}), and every $t$ reachable from a state of $B$ belongs to~$B$. 

We assume familiarity with basic results of ergodic theory and calculus that are recalled at appropriate places.

\subsection{Terrain model}
\label{sec-terrain}

The terrain is modeled as a finite directed graph $G = (V,E,\tm)$ where the vertices of $V$ correspond to robot's positions, the edges $E \subseteq V \times V$ are possible moves, and $\tm \colon E\to\Nset_+$ specifies the traversal time of an edge. Later, we adjoin additional parameters to edges and vertices modeling the costs, importance, etc.

We require that $G$ is strongly connected, \ie, for all $v,u \in V$ there is a path from $v$ to $u$. We write $u\to v$ instead of $(u,v)\in E$. 

\subsection{Moving strategy}
\label{sec-Defender-strategy}

Let us fix a graph $G = (V,E,\tm)$, and let $\Mem$ be a non-empty set of \emph{memory states}. Intuitively, memory states are used to ``remember'' some information about the history of visited vertices. 
Since infinite-memory is not implementable, from now on we restrict ourselves to \emph{finite-memory} strategies. 

A robot's \emph{configuration} $(v,m)$ is determined by the currently visited vertex $v$ and the current memory state $m$. We use $\conf = V \times \Mem$ to denote the set of all configurations. A configuration of the form $(v,m)$ is written as $\widehat{v}$.

When the robot visits a vertex, the next move is chosen randomly according to the current configuration. Formally, a \emph{moving strategy} for $G$ with memory $\Mem$ is a function \mbox{$\sigma\colon \conf \to \Dist(\conf)$} such that $\sigma(\widehat{v})(\widehat{u}) > 0$ only if $v \to u$. If $\sigma(\widehat{v})$ is a Dirac distribution for every $\widehat{v} \in \conf$, we say that $\sigma$ is \emph{deterministic}.

Every moving strategy $\sigma$ determines a Markov chain $G_\sigma$ where $\conf$ is the set of states and $P(\widehat{v},\widehat{u}) = \sigma(\widehat{v})(\widehat{u})$. 
The \emph{edges} of $G_\sigma$ are defined by $\widehat{v} \to \widehat{u}$  iff $v \to u$. Note that some edges may have zero probability, \ie, $\sigma(\widehat{v})(\widehat{u}) = 0$, but the BSCCs of $G_\sigma$ are determined by the edges with positive probability (see the above definition of Markov chain).  
The \emph{traversal time} of an edge $\widehat{v} \to \widehat{u}$ is the same as in $G$, \ie, equal to $\tm(v,u)$. 

\subsection{Recurrent reachability objectives}
For the rest of this section, we fix a graph $G = (V,E,\tm)$ and a finite set $\Mem$ of memory states.

\subsubsection{Atomic expressions}
We start by introducing basic expressions used to construct recurrent reachability objectives.

A \emph{walk} in $G_\sigma$ is an infinite sequence $w = \widehat{v}_0,\widehat{v}_1,\ldots$ such that $\widehat{v}_i \to \widehat{v}_{i+1}$ for all $i \in \Nset$. Let $C \subseteq \conf$.
We say that $w$ \emph{hits $C$ in time~$t$} if $t = \sum_{i=0}^{m-1} \tm(v_i,v_{i+1})$, where $m$ is the least index $j$ such that $\widehat{v}_j \in C$. If there is no such $j$, we put $t = \infty$.
For all $\widehat{v} \in \conf$ and $C \subseteq \conf$, we use
$\Tm(\widehat{v} {\to} C)$ to denote the expected hitting time of $C$ by a walk initiated in $\widehat{v}$ (if $\widehat{v} \in C$, then $\Tm(\widehat{v} {\to} C) = 0$), and
$\Tm^2(\widehat{v} {\to} C)$ to denote the expected \emph{square} of the hitting time of $C$ by a walk initiated in $\widehat{v}$.
Furthermore, for every walk $w = \widehat{v}_0,\widehat{v}_1,\ldots$ and every edge $\widehat{e} = \widehat{v} \to \widehat{u}$, we define the \emph{frequency of $\widehat{e}$ along $w$} as the limit percentage of time spent by executing $\widehat{e}$ along $w$, \ie,
\begin{equation}
	\small
	\F(\widehat{e},w) = \lim_{m \to \infty} \frac{\tm(v,u) \cdot \#_{\widehat{e}}(\widehat{v}_0,\ldots,\widehat{v}_{m+1})}{\sum_{i=0}^m \tm(v_i,v_{i+1})},
\end{equation}
where $\#_{\widehat{e}}(\widehat{v}_0,\ldots,\widehat{v}_{m+1})$ is the number of occurrences of $\widehat{e}$ in the prefix $\widehat{v}_0,\ldots,\widehat{v}_{m+1}$. It follows from basic results of Markov chain theory that the above limit is defined for \emph{almost all} walks (\ie, with probability one), and almost all walks $w$ that hit the same BSCC of $G_\sigma$ have the same $\F(\hat{e},w)$.

\subsubsection{Syntax}
Recurrent reachability objective functions are closed-form expressions over numerical constants and atomic expressions of the form $\Tm(\widehat{v} {\to} C)$, $\Tm^2(\widehat{v} {\to} C)$, $\F(\widehat{e})$, and $p(\dhat{e})$ obtained by using 
\begin{itemize}
\item addition, multiplication, min, and max that may take arbitrarily many arguments;
\item division, where the denominator is an expression over numerical constants and atomic expressions of the form $\F(\widehat{e})$, $p(\widehat{e})$  built using addition and multiplication. 
\item other differentiable functions such as square root that are defined for all non-negative arguments.
\end{itemize}
When defining the arguments of sums, products, min, and max, we may refer to special sets $\Ac$ and $\Ae$ consisting of \emph{active} configurations and edges, respectively, whose semantics is defined in the next paragraph. 

A \emph{recurrent reachability optimization problem} is a problem of the form $\textbf{minimize}~R$ or $\textbf{maximize}~R$, where $R$ is a recurrent reachability objective function.

\subsubsection{Ergodicity}

Infinite-horizon objective functions are typically independent of finite prefixes of runs and the initial configuration can be chosen freely.
Hence, the objective value for $G_\sigma$ actually depends only on the objective values attained in the ``best'' BSCC of $G_\sigma$.
From now on, we only consider recurrent reachability objective functions $R$ satisfying this condition (this is equivalent to requiring that $R$ can be optimized by an \emph{ergodic} strategy where $G_\sigma$ is strongly connected).

\subsubsection{Evaluation}

Let $\sigma$ be a moving strategy for $G$ with memory $\Mem$, and let $B$ be a BSCC of $G_\sigma$. For a given recurrent reachability function $R$, we use $R[B]$ to denote the \emph{value of $R$ in $B$}, defined by structural induction as follows:
\begin{itemize}
\item atomic expressions $\Tm(\widehat{v} {\to} C)$, $\Tm^2(\widehat{v} {\to} C)$, $\F(\hat{e})$, and $p(\dhat{e})$ are evaluated in the way described above ($p(\dhat{e})$ is the probability of $\dhat{e}$ assigned by $\sigma$).
In particular, note that $\F(\widehat{e})$ where $\widehat{e} = \widehat{v} {\to} \widehat{u}$ can be positive only if $\widehat{v},\widehat{u} \in B$.
\item The set $\Ac$ of active configurations is equal to $B$, and the set $\Ae$ of active edges consists of all $(\widehat{v},\widehat{u})$ such that  $\widehat{v},\widehat{u} \in B$ and $\sigma(\widehat{v})(\widehat{u}) > 0$.
\item The addition, multiplication, min and max are evaluated in the expected way. If the set of arguments is parametrized by $\Ac$ or $\Ae$, it is constructed for the set of configurations and edges defined in the previous item. 
\end{itemize}
In some cases, $R[B]$ can be \emph{undefined} (see Section~\ref{sec-algo}). 

Finally, we define the \emph{$\sigma$-value} of the objective $\textbf{minimize}~R$ or $\textbf{maximize}~R$ as the minimal or the maximal $R[B]$ such that $B$ is a BSCC of $G_\sigma$ where $R[B]$ is defined (if there is no such $B$, then the $\sigma$-value in undefined).

\section{Examples}
\label{sec-examples}

To demonstrate the versatility of recurrent reachability objectives, we present selected examples of concrete objective functions. The list is \emph{by no means exhaustive}---we show how to capture some of the existing infinite-horizon optimization criteria and how to extend them to control various forms of stochastic instability caused by randomization. Let us emphasize that our aim is to illustrate the \emph{expressive power} of our optimization framework, not to design the most appropriate objectives capturing the discussed phenomena.

For this section, we fix a graph $G = (V,E,\tm)$ and a finite set $\Mem$ of memory states. For a given subset $U \subseteq V$, we use $C^U$ to denote the subset $U \times M$ of configurations.

For simplicity, in the first two subsections we  assume that the traversal time of every edge is~$1$.

\subsubsection{Mean Payoff}
As a warm-up, consider the concept of \emph{mean payoff}, \ie, the long-run average payoff per visited vertex. Let $\alpha\colon V \rightarrow \Nset_+$ be a \emph{payoff function}.
The goal is to minimize the mean payoff for $\alpha$. This is formalized as \textbf{minimize}~\text{MP}, where
\begin{align}
\small
    \textrm{MP} &\equiv \sum_{\widehat{v} \in \conf} F(\widehat{v}) \cdot \alpha(v),
		\label{eq:MP}
		\\
 F(\widehat{v}) &\equiv \sum_{\widehat{e} \in \Out(\widehat{v})} \F(\widehat{e}) \label{eq-freq}.
\end{align}
Here, $\Out(\widehat{v})$ is the set of all out-going edges of $\widehat{v}$. Hence, $F(\widehat{v})$ is the limit frequency of visits to $\widehat{v}$. 

Minimizing mean-payoff is computationally easy. However, the costs of vertices visited by an optimal strategy can significantly differ from the mean payoff (\ie, the costs are not distributed sufficiently ``smoothly'' along a walk). If ``smoothness'' is important, it can be enforced, \eg, by the objective  
$\textbf{minimize}~\text{MP} + \beta \cdot \text{DMP}$, where $\beta$ is a suitable weight and
\begin{equation} \label{eq:DMP}
	\small
	\textrm{DMP}
		\equiv \sqrt{\sum_{\widehat{v} \in \conf} F(\widehat{v}) \cdot \bigl(\text{MP} -\alpha(v)\bigr)^2}
\end{equation}
is the standard deviation of the costs per visited vertex. Similarly, we can formalize objectives involving multiple cost/payoff functions and enforce some form of ``combined smoothness'' if appropriate.

\subsubsection{Renewal Time}
Let $T \subseteq V$ be a set of targets, and consider the problem of minimizing the maximal renewal time of a target. First, for every $\widehat{\tau} \in C^T$, let 
\begin{equation}
	P(\widehat{\tau}) \equiv  \frac{F(\widehat{\tau})}{\sum_{\kappa \in C^{\tau}} F(\kappa)},
\end{equation}
where $F(\cdot)$ is defined by~\eqref{eq-freq}. Hence, $P(\widehat{\tau})$ is the percentage of visits to $\widehat{\tau}$ among all visits to configurations of $C^{\tau}$, assuming that the denominator is positive. Then, the renewal time of $\tau$, denoted by $\text{ERen}(\tau)$ is given
by 
\begin{equation}
\label{eq:ERen}
 \sum_{\widehat{\tau} \in C^\tau} P(\widehat{\tau})
	 \sum_{\dhat{\tau} {\to}\dhat{u}}p(\dhat{\tau},\dhat{u})
	 \cdot \bigl(1+ \Tm(\widehat{u} {\to} C^\tau)\bigr)
\end{equation}
and the expected \emph{square} of the time needed to revisit $\tau$, denoted by $\text{QRen}(\tau)$, is expressible as
\begin{equation}
  \sum_{\widehat{\tau} \in C^\tau}\! P(\widehat{\tau})
		\!\sum_{\dhat{\tau} {\to}\dhat{u}}p(\dhat{\tau},\dhat{u})
		\!\cdot\!\bigl(1 + 2\Tm(\widehat{u} {\to} C^\tau) + \Tm^2(\widehat{u} {\to} C^\tau)\bigr).
\end{equation}
Minimizing the maximal renewal time is then formalized as  $\textbf{minimize}\ \max_{\tau \in T} \text{ERen}(\tau)$. However, this simple objective does not take into account possible deviations of the renewal time from its mean. This can be captured, \eg, by expressing the standard deviation as
\begin{equation}
\label{eq:DevRen}
 \text{DevRen}(\tau) \equiv \sqrt{\text{QRen}(\tau) - (\text{ERen}(\tau))^2}
\end{equation}
and using $\textbf{minimize}\ \max_{\tau \in T} (\text{ERen}(\tau) + \beta\cdot\text{DevRen}(\tau))$ for a suitable weight~$\beta$.

\subsubsection{Patrolling} 
Let $T \subseteq V$ be a set of targets, and $\alpha\colon T {\to} \Nset_+$ a function assigning to every target its \emph{importance}. The \emph{damage} caused by an attack at a target $\tau$ (such as setting a fire) is given as $\alpha(\tau) \cdot t$ where $t$ is the time to discover the attack by the robot. The patrolling objective is to minimize the expected damage.

Patrolling problems are studied for \emph{adversarial} and \emph{non-adversarial} environments. In the first case, there is an active Attacker knowing the robot's strategy and observing its moves. For the Attacker, an appropriate moment to initiate an attack is when the robot leaves a vertex and starts walking along some edge $\widehat{v} {\to} \widehat{u}$. The objective is to minimize the expected damage over all possible attacks, \ie,  
\begin{equation} \label{eq-adversarial}
  \textbf{minimize}
		\ \max_{\widehat{v}{\to}\widehat{u}\in \Ae}
		\ \max_{\tau \in T}\ \alpha(\tau)
				\cdot \bigl(\tm(v,u) + \Tm(\widehat{u} {\to} C^\tau)\bigr).
\end{equation}
Note that~\eqref{eq-adversarial} conveniently uses the set $\Ae$ of active edges to restrict the $\max$ only to ``relevant'' edges used by the strategy. 

In non-adversarial patrolling, the attacks are performed by ``nature''. Let $\pi$ be a distribution over $T$ specifying the attack chance (such as the probability of spontaneous ignition). Then, minimizing the expected damage is expressible as \mbox{\textbf{minimize}\ \text{EDam}}, where the function \text{EDam} is defined as
\begin{equation} \label{eq-nonadversarial}
	\small
	\sum_{\widehat{v}{\to}\widehat{u}\in \Ae} \F(\dhat{v},\dhat{u})
		\sum_{\tau \in T} \pi(\tau) \alpha(\tau)
		\bigg(\frac{\tm(v,u)}{2} + \Tm(\widehat{u} {\to} C^\tau)\bigg).
\end{equation}
Note that if $\tau$ is attacked when the robot walks along $\widehat{v}{\to}\widehat{u}$, then the robot is in the middle of this edge \emph{on average}. Hence, the average time to reach $\tau$ is $\tm(v,u)/2 + \Tm(\widehat{u} {\to} C^\tau)$.

Again, we can express the variances/deviations of the relevant random variables (incl.{} the variance of the expected damage of~\eqref{eq-nonadversarial}). These expressions are relatively long, but their construction is straightforward.

\section{The Algorithm}
\label{sec-algo}

The algorithm is based on gradient descent: It starts with a random initial strategy and then repeatedly evaluates the objective function $R$ and modifies the current strategy in the direction of the gradient $\nabla R$ (or $-\nabla R$ for minimization).
After a number of iterations, the strategy attaining the best objective value is returned.

Let $\sigma$ be a moving strategy for a graph $G$. We show how to evaluate and differentiate $R[B]$ for a given BSCC $B$ of~$G_\sigma$. The atomic expressions
are obtained as unique solutions of linear equations systems. More concretely, for a target set $C$, active configurations $\Ac$ and edges $\Ae$, we have that $\Tm(\dhat{v} {\to} C)$ and $\Tm^2(\dhat{v} {\to} C)$ are \emph{undefined} for $\dhat{v} \not\in \Ac$. If $\dhat{v} \in \Ac$ and $\Ac \cap C = \emptyset$, then 
$\Tm(\dhat{v} {\to} C) = \Tm^2(\dhat{v} {\to} C) = \infty$. If $\Ac \cap C \neq \emptyset$, then for every $\dhat{v} \in \Ac$ we fix a variable $X_{\dhat{v}}$ and an equation
\begin{equation*} \label{E:system}
	X_{\dhat{v}}  = 
  	\begin{cases}
  		0 & \text{if $\dhat{v} \in C$,}\\
			\sum_{\dhat{v} \to \dhat{w}} \sigma(\dhat{v})(\dhat{w}) \cdot \bigl(\tm(v,w)+X_{\dhat{w}}\bigr)
				& \text{otherwise.}
		\end{cases}
\end{equation*}
Then, the tuple of all $\Tm(\dhat{v} {\to} C)$, where $\dhat{v} \in \Ac$, is the unique solution of this system. 

Similarly, if $\Ac \cap C \neq \emptyset$, then the tuple of all $\Tm^2(\dhat{v} {\to} C)$, where $\dhat{v} \in \Ac$, is the unique solution of the system where to each $\dhat{v}\in \Ac$, we assign a variable $Y_{\dhat{v}}$ and an equation
\begin{equation*} \label{E:system2}
	Y_{\dhat{v}}  =
		\begin{cases}
			0 & \text{if $\dhat{v} \in C$,}\\
				\sum_{\dhat{v} \to \dhat{w}} \sigma(\dhat{v})(\dhat{w}) \cdot \gamma(\dhat{v},\dhat{w})
				& \text{otherwise,}
		\end{cases}
\end{equation*}
where $\gamma(\dhat{v},\dhat{w})=\tm(v,w)\bigl(\tm(v,w)+2\Tm(\dhat{w} {\to} C)\bigr)+Y_{\dhat{w}}$.

To compute the edge frequencies, we fix a variable $Z_{\dhat{v}}$ for every $\dhat{v} \in \Ac$,  and an equation
\begin{equation*} \label{E:system3}
	Z_{\dhat{v}} 
		= \textstyle\sum_{\dhat{w} \to \dhat{v}} \sigma(\dhat{w})(\dhat{v}) \cdot Z_{\dhat{w}}.
\end{equation*}
This system, together with the equation $\sum_{\dhat{v}\in \Ac}Z_{\dhat{v}}=1$, has a unique solution $\F$ where $\F(\dhat{v})$ is the frequency of visits to~$\dhat{v}$. For each edge $\dhat{e}=(\dhat{v},\dhat{w})\in\Ae$, we set $D(\dhat{e})= \F(\dhat{v})\cdot\sigma(\dhat{v})(\dhat{w})\cdot\tm(v,w)$
and we get
\begin{equation*}
	\F(\dhat{e}) = {D(\dhat e)}/{\textstyle\sum_{\dhat \epsilon\,\in\Ae} D(\hat \epsilon\,)}.
\end{equation*}
For the other edges $\dhat{e} \not\in \Ae$, we have that $\F(\dhat{e}) = 0$. The value of $R[B]$ is then obtained from the atomic expressions in the straightforward way (when some atomic expression used in $R$ is undefined or the denominator of some fraction of $R$ is zero in $B$, then $R[B]$ is undefined).

Next, we need to calculate the gradient $\nabla R$.
As objectives are, by design, allowed to use only smooth operations with $\min$ and $\max$ over atomic expressions,
the derivatives of $R$ \wrt these atomic expressions are well defined almost everywhere.
Solutions of systems of linear equations depend smoothly on the parameters and the derivatives of our atomic expressions \wrt $\sigma$ can be calculated as solutions of other linear systems. 
We use PyTorch Library~\cite{PyTorch} and its automatic differentiation in our implementation.

However, a na\"ive update $\sigma\pm\lambda\nabla R$ for a step size $\lambda$ almost never yields a probability distribution (\ie, a valid strategy).
The standard approach is to produce strategies from real-valued coefficients by a \textbf{Softmax} function. 
Any update of these coefficients then leads to a well-defined strategy.
The drawback of \textbf{Softmax} is that the resulting distributions never contain zeros (\ie, the strategies always use all edges).

To reach all possible strategies, we cut the small probabilities at a certain threshold (and normalize) by \textbf{Cutoff} function.
However, as edges with zero probabilities are excluded from $\Ae$, discontinuities in the objective may occur.
For instance, in Patrolling objective~\eqref{eq-adversarial}, the term $\tm(v,u)$ is present for an edge $\dhat v{\to}\dhat u$ 
if $\sigma(\dhat v)(\dhat u)>0$ and drops to zero if $\sigma(\dhat v)(\dhat u)=0$.

In other words, objective $R$ as a function of real-valued coefficients is a smooth function on an open set with possible jumps at the boundary. 
In order to make the boundary values accessible by the gradient descent, we relax our discontinuous $R$ to a smooth one, say $R^*$.
For instance, in Patrolling objective~\eqref{eq-adversarial}, we multiply $\tm(u,u)$ by a factor $\operatorname{\textbf{HardTanh}}(\varepsilon\sigma(\dhat v, \dhat u))$, which interpolates the values 0 and 1 continuously.
Moreover, for a more efficient gradient propagation, we replace each $\min$ and $\max$ with their relaxed variants as in \cite{kla:18}.

The final algorithm is described in Procedure~\ref{alg:optim}.
Strategy coefficients are initialized at random.
In every step, we compute the relaxed objective $R^*$ value of the current strategy and update the coefficients based on the gradient $\nabla R^*$ using Adam optimizer~\cite{Adam}.
We also add a decaying Gaussian noise to the gradient.
Then, we round the strategy using \textbf{Cutoff} and compute its true objective value $R$.
Procedure~\ref{alg:optim} can be run repeatedly to increase the chance of producing a strategy with better value.

\makeatletter
\renewcommand{\ALG@name}{Procedure}
\makeatother

\begin{algorithm}
\small
\caption{Strategy optimization}
\label{alg:optim}
\begin{algorithmic}
\State coefficients $\gets$ \textbf{Init}()
\For{step $\in$ steps} 
	\State strategy $\gets$ \textbf{Softmax}(coefficients)
	\LineComment{Evaluating relaxed objective and its gradient}
	\State value $\gets$ \textbf{Eval}($R^*$, strategy)
	\State coefficients.grad $\gets$ \textbf{Gradient}(value)
	\LineComment{Gaussian noise and Adam optimizer's step}
	\State coefficients.grad += \textbf{Noise}(step)
	\State coefficients += \textbf{Step}(coefficients.grad, step)
	\LineComment{Strategy evaluation}
	\State strategy $\gets$ \textbf{Cutoff}(\textbf{Softmax}(coefficients))
	\State value $\gets$ \textbf{Eval}($R$, strategy)
	\State \textbf{Save} value, strategy\vspace{0.5ex}
\EndFor
\Return strategy with the lowest/highest value
\end{algorithmic}
\end{algorithm}

The algorithm is efficient because it does not involve any computationally demanding tasks, but it does not guarantee (sub)optimality of the constructed strategies.
This is unavoidable due to the $\mathsf{NP}$-hardness of some of the recurrent reachability objectives.

\section{Experiments}
\label{sec-experiments}

There is no previous work setting the baseline for evaluating the quality of strategies produced by our algorithm.
Therefore, we selected two representative examples where the tradeoffs between performance and stability are not easy to discover, but the functionality of the synthesized strategies can still be analyzed by hand and compared against the best \emph{achievable} outcomes identified by theoretical analysis.%
\footnote{The code for reproducing the results is available at \url{https://gitlab.fi.muni.cz/formela/2022-ijcai-optimization-framework}.
See also the latest version at \url{https://gitlab.fi.muni.cz/formela/regstar}.}

\subsubsection{Mean Payoff}
\begin{figure}[h]\centering
\begin{tikzpicture}[x=1.7cm, y=2.3cm, scale=0.5,font=\small]
    \node [min] (V1) at (0,0) {$7$};
    \node [min] (V2) at (1,0) {$0$};
    \node [min] (V3) at (2,0) {$6$};
    \node [min] (V4) at (3,0) {$5$};
    \node [min] (V5) at (4,0) {$7$};
    \draw[-stealth, thick] (V1) -- (V2);
    \draw[-stealth, thick] (V3) -- (V4);
    \draw[-stealth, thick] (V4) -- (V5);
    \draw[-stealth, thick, rounded corners] (V1) -- ($(V1) +(.3,.4)$) -- ($(V3) +(-.3,.4)$) -- (V3);
    \draw[-stealth, thick, rounded corners] (V2) -- ($(V2) +(.3,-.4)$) -- ($(V4) +(-.3,-.4)$) -- (V4);
    \draw[-stealth, thick, rounded corners] (V3) -- ($(V3) +(.3,.4)$) -- ($(V5) +(-.3,.4)$) -- (V5);
    \draw[-stealth, thick, rounded corners] (V5) -- ($(V5) +(-.3,-.6)$) -- ($(V1) +(.3,-.6)$) -- (V1);
		\path (4.5,-0.1) node[right]{
			\small
			\begin{tabular}{ccc}
			Cycle & MP & DMP \\
			\toprule
			7-6-7   & $6.667$ & $  0.471$\\
			7-6-5-7 & $6.250$ & $  0.829$\\
			7-0-5-7 & $4.750$ & $  2.861$
			\end{tabular}};
\end{tikzpicture}
\caption{Optimization of $\text{MP}+\beta\cdot\text{DMP}$ results in choosing one of the three cycles according to the chosen $\beta$. Costs of the targets are displayed; all edges are of unit length.}
\label{fig-mp}
\end{figure}
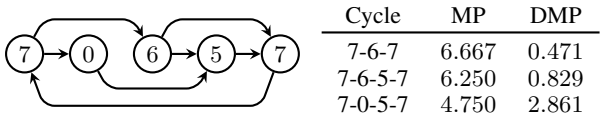

We minimize the objective $\text{MP} + \beta \cdot \text{DMP}$ defined in \eqref{eq:MP} and \eqref{eq:DMP} for the graph of Fig.~\ref{fig-mp}.
The graph contains three cycles with the corresponding MP and DMP as in the table of Fig.~\ref{fig-mp}.

For almost every $\beta$, the objective's minimizer is \emph{precisely one} of the three cycles.
More precisely, it is the cycle 7-0-5-7 for all $\beta \in [0,\varrho_1]$, the cycle 7-6-5-7 for all  $\beta \in [\varrho_1,\varrho_2]$, and the cycle 7-6-7 for all $\beta \in [\varrho_2,\infty)$, where $\varrho_1 \approx 0.73$ and $\varrho_2 \approx 1.16$.
Ideally, our algorithm should find the corresponding cycle for every $\beta$.

\begin{figure}[h]\centering
\includegraphics[width=\linewidth]{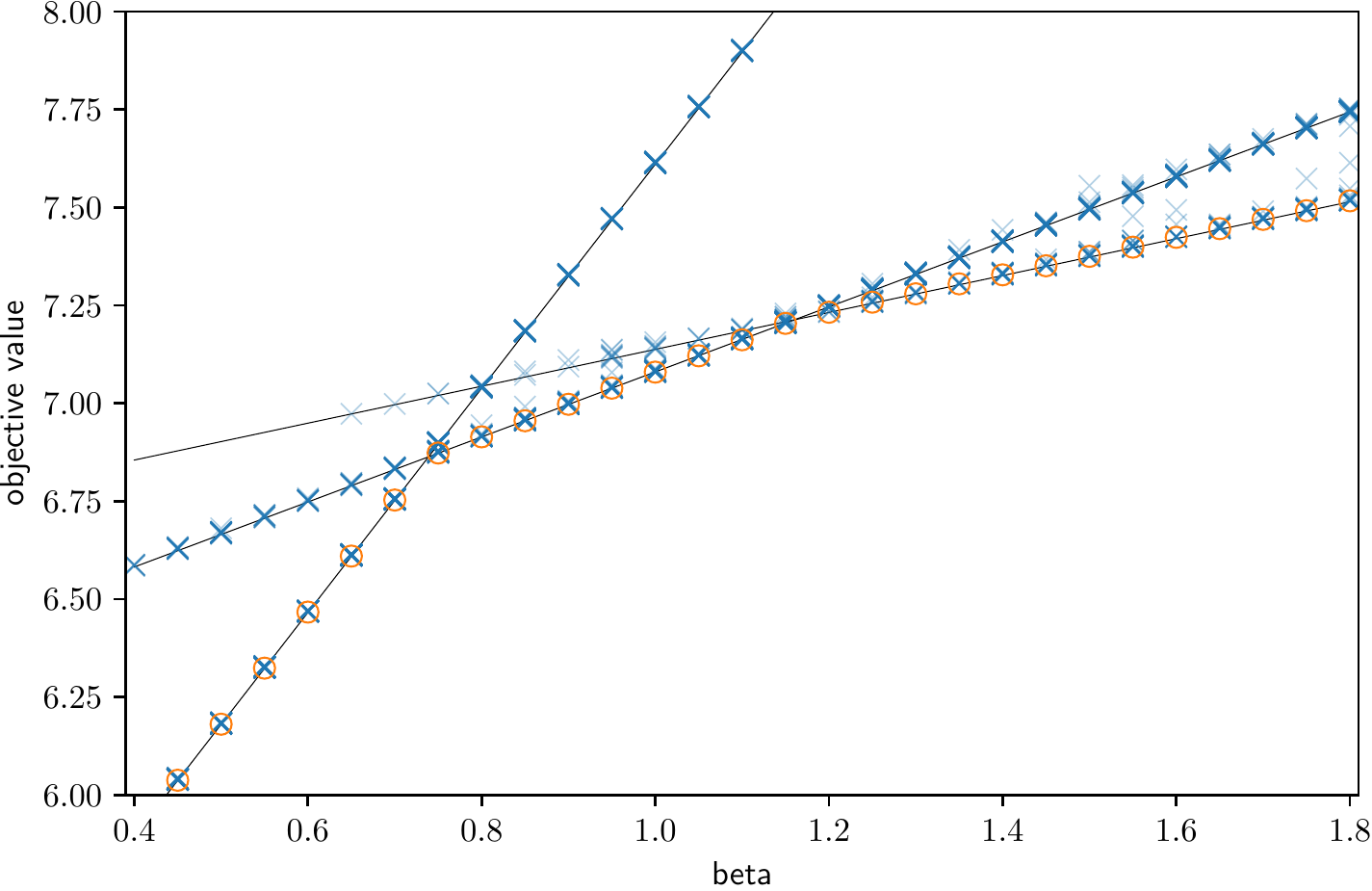}
\caption{The best values of Mean Payoff and its standard deviation $\text{MP} + \beta \cdot \text{DMP}$ for different values of $\beta$. The three lines correspond to the three strategies discussed in Fig.~\ref{fig-mp}. The best-obtained values among 100 trials (circled) fit the optima.}
\label{fig-MP-values}
\end{figure}

The algorithm outcomes are shown in Fig.~\ref{fig-MP-values}. For every $\beta$, we perform 100 trials (\ie, construct 100 strategies with one memory state), and report the corresponding $\text{MP} + \beta \cdot \text{DMP}$ value. The value of the best strategy achieved for a given $\beta$ is represented by a ``circle''; the other ``crosses'' represent the values of non-optimal strategies. Observe that the circles agree with the ideal outcomes. 

\subsubsection{Renewal time}

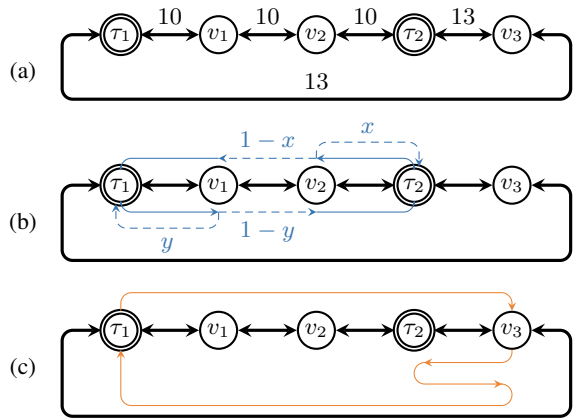
\begin{figure}[h]\centering
\begin{tikzpicture}[x=2cm, y=1.1cm, scale=0.65,font=\small]
\foreach \x/\c/\l in {0/2.7/b,1/0/c,2/5.5/a}{%
    \coordinate (a) at (0,\c);
    \node at ($(a) +(-1,-0.7)$) {(\l)};
    \node [min,double] (T1\x) at (a) {$\tau_1$};
    \node [min,double] (T2\x) at ($ (a) + (3,0) $) {$\tau_2$};
    \node [min] (V1\x) at ($ (a) + (1,0) $) {$v_1$};
    \node [min] (V2\x) at ($ (a) + (2,0) $) {$v_2$};
    \node [min] (V3\x) at ($ (a) + (4,0) $) {$v_3$};
    \ifthenelse{\x=2}{%
    \draw[stealth-stealth,very thick] (T1\x) -- node[above] {$10$}  (V1\x);
    \draw[stealth-stealth,very thick] (V1\x) -- node[above] {$10$} (V2\x);
    \draw[stealth-stealth,very thick] (V2\x) -- node[above] {$10$} (T2\x);
    \draw[stealth-stealth,very thick] (T2\x) -- node[above] {$13$} (V3\x);
    \draw[stealth-stealth,very thick,rounded corners] (V3\x) -- ($(V3\x) +(.6,0)$) -- ($(V3\x) +(.6,-1.2)$) --  node[above] {$13$}    ($(T1\x) +(-.6,-1.2)$)   |-  (T1\x);
    }{%
    \draw[stealth-stealth,very thick] (T1\x) -- (V1\x);
    \draw[stealth-stealth,very thick] (V1\x) -- (V2\x);
    \draw[stealth-stealth,very thick] (V2\x) -- (T2\x);
    \draw[stealth-stealth,very thick] (T2\x) --  (V3\x);
    }
    \ifthenelse{\x=1}{%
    \draw[-stealth,rounded corners,color=ourorange] (T1\x) -- ($(T1\x) +(0,.7)$)  -| (V3\x); 
    \draw[-stealth,rounded corners,color=ourorange] (V3\x) |- ($(V3\x)!0.9!(T2\x) + (0,-.6)$);
    \draw[-stealth,rounded corners,color=ourorange] ($(V3\x)!0.9!(T2\x) - (0,.6)$) -| ($(T2\x) - (0,1)$) -- ($(V3\x)!0.1!(T2\x) - (0,1)$);
    \draw[-stealth,rounded corners,color=ourorange] ($(V3\x)!0.1!(T2\x) + (0,-1)$) -| ($(V3\x) +(0,-1.4)$)  -| (T1\x);
    \draw[stealth-stealth,very thick,rounded corners] (V3\x) -- ($(V3\x) +(.6,0)$) -- ($(V3\x) +(.6,-1.6)$) -- ($(T1\x) +(-.6,-1.6)$)   |-  (T1\x);    
    }{} 
    \ifthenelse{\x=0}{%
    \draw[-stealth,rounded corners, color=ourblue] (T1\x) -- ($(T1\x) +(0,-.5)$) -- ($(V1\x) + (0,-.5)$);  
    \draw[rounded corners, color=ourblue] ($(V1\x) +(0,.5)$) -| (T1\x);  
    \draw[-stealth,rounded corners, color=ourblue] (T2\x) -- ($(T2\x) +(0,.5)$) -- ($(V2\x) + (0,.5)$);
    \draw[rounded corners, color=ourblue] ($(V2\x) +(0,-.5)$) -| (T2\x);  
    \draw[-stealth, densely dashed, rounded corners, color=ourblue] ($(V2\x) +(0,.5)$) -- node[above]{$1-x$} ($(V1\x) +(0,.5)$); 
    \draw[-stealth, densely dashed, rounded corners, color=ourblue] ($(V1\x) +(0,-.5)$) -- node[below]{$1-y$} ($(V2\x) +(0,-.5)$); 
    \draw[-stealth, densely dashed, rounded corners, color=ourblue] ($(V2\x) +(0,.5)$) -- ($(V2\x) +(0,.8)$) --
       node[above]{$x$}  ($(T2\x) +(.05,.8)$) -- ($(T2\x) + (0.05,0.35)$);    
    \draw[-stealth, densely dashed, rounded corners, color=ourblue] ($(V1\x) +(0,-.5)$) -- ($(V1\x) +(0,-.8)$) --
       node[below]{$y$}  ($(T1\x) +(-.05,-.8)$) -- ($(T1\x) - (0.05,0.35)$);  
    \draw[stealth-stealth,very thick,rounded corners] (V3\x) -- ($(V3\x) +(.6,0)$) -- ($(V3\x) +(.6,-1.4)$) -- ($(T1\x) +(-.6,-1.4)$)   |-  (T1\x);
    }{}
}
\end{tikzpicture}
\caption{Minimizing the max.~expected renewal time and its std.~deviation $\text{ERen}+\beta\cdot\text{DevRen}$ in graph (a). Randomized strategy (b) is optimal for small $\beta$ and deterministic loop (c) prevails for $\beta$~large.}
\label{fig-renewal}
\end{figure}

\begin{figure}[h!]\centering
\includegraphics[width=\linewidth]{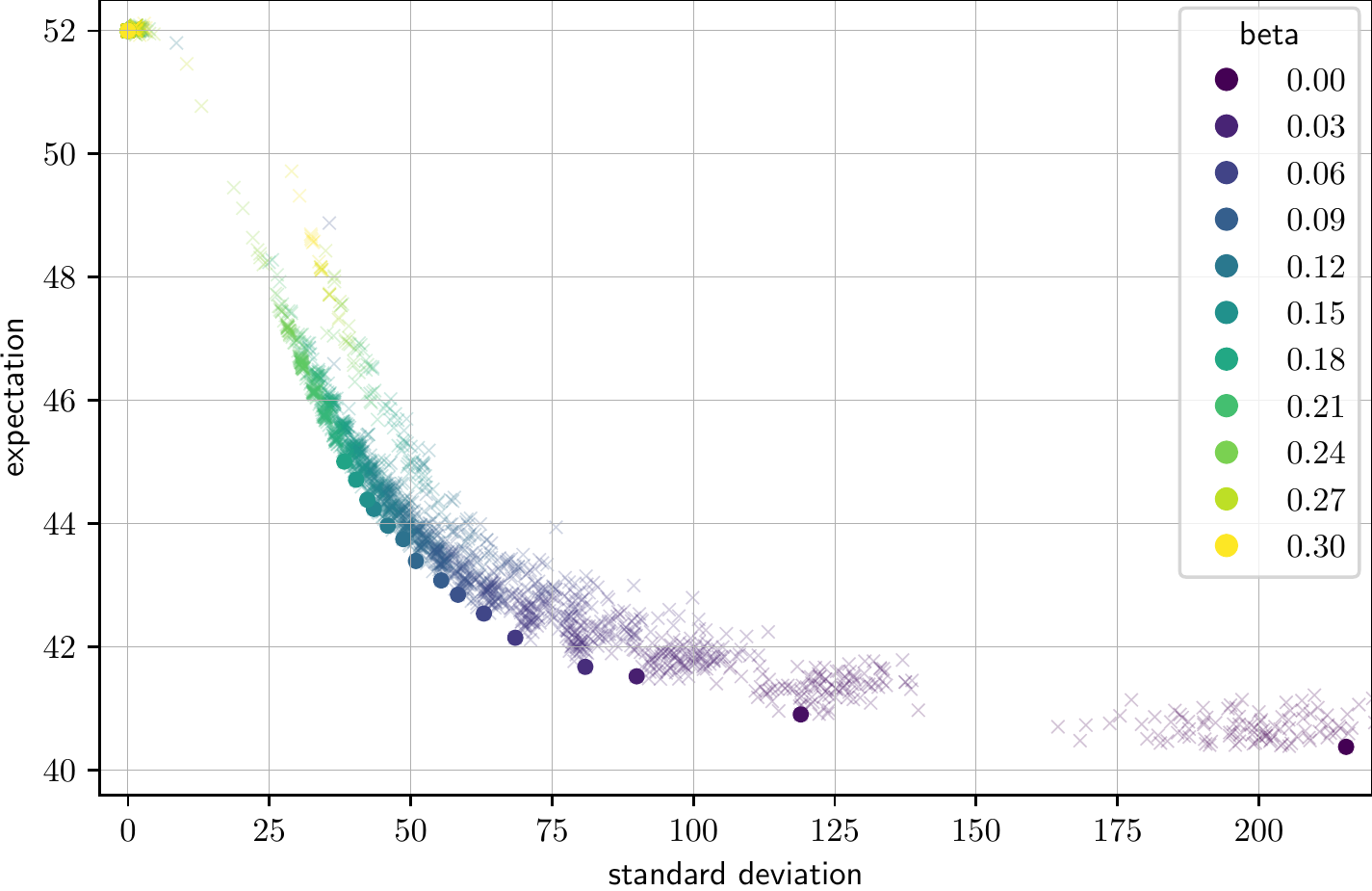}
\caption{Maximal expected renewal time of a target and its std.~deviation $\text{ERen}+\beta\cdot\text{DevRen}$ obtained for 100 trials with various $\beta$.}
\label{fig-cinka-pareto}
\end{figure}

Consider the graph of Fig.~\ref{fig-renewal}a.
We minimize the objective $\max_{\tau \in T} \bigl(\text{ERen}(\tau) + \beta\cdot\text{DevRen}(\tau)\bigr)$ defined in \eqref{eq:ERen} and \eqref{eq:DevRen}.
The outcomes of our algorithm for two memory states are shown in Fig.~\ref{fig-cinka-pareto}.
For each $\beta$ ranging from $0$ to $0.3$, we run 100 trials and report the expected renewal time and the corresponding standard deviation of the resulting 100 strategies;
the best values are highlighted by solid dots.
The values of the obtained strategies are concentrated near the best one, showing the stability of the optimization.

For smaller $\beta$, the constructed strategies have a smaller renewal time but large deviation, and they work as shown in  Fig.~\ref{fig-renewal}b.
That is, they tend to re-visit $\tau_1$ from $v_1$ and $\tau_2$ from $v_2$.
As $x,y \rightarrow 1$, the maximal renewal time approaches $40$ and the standard deviation approaches $\infty$.

For larger $\beta$, where the standard deviation is punished more severely, the algorithm tends to decrease $x$, $y$.
Note that for $x=y=0$, the strategy of Fig.~\ref{fig-renewal}b becomes deterministic with renewal time $60$ and zero deviation.
However, this point is \emph{not reached} by the curve of Fig.~\ref{fig-cinka-pareto}.
The reason is that for $\beta \approx 0.27$, the algorithm discovers and strongly prefers a \emph{completely different strategy}, which goes through $v_3$ instead (see Fig.~\ref{fig-renewal}c), with maximal renewal time $52$ and zero deviation (this is the \emph{best} strategy with zero deviation).

\section{Conclusions}
\label{sec-concl}

The obtained optimization framework for recurrent reachability objectives is applicable not only to graphs, but also to Markov decision processes without any additional effort.
Here, the randomization introduced by the strategies is combined with the ``inherent randomization'' of the model.
Here, stochastic instability cannot be removed completely, and a precise understanding of the principal limits of this effort is a challenging direction for future research.
 
\clearpage
\section*{Acknowledgements}
This work is supported by the Czech Science Foundation, Grant No.~21-24711S,
and from Operational Programme Research, Development and
Education -- Project Postdoc2MUNI No.\ CZ.02.2.69/0.0/0.0/18\_053/0016952.

\end{document}